\documentclass[11pt]{article}

\usepackage[utf8]{inputenc}
\usepackage[T1]{fontenc}
\usepackage{lmodern}
\usepackage{microtype}
\usepackage[margin=1in]{geometry}
\usepackage{booktabs}
\usepackage{longtable}
\usepackage{amsmath,amssymb}
\usepackage{array}
\usepackage{authblk}
\usepackage{xurl}
\usepackage[colorlinks=true,linkcolor=blue,citecolor=blue,urlcolor=blue]{hyperref}

\title{Evaluating Prompt Scope and Demonstration Similarity in Local LLM Machine Translation}

\author[1]{Mihael Arcan}
\affil[1]{Home Lab, Galway, Ireland}

\date{}

\begin{document}
\maketitle

\begin{abstract}
Large language models (LLMs) are increasingly used as general-purpose translation systems, but their behavior is usually evaluated under a single prompt shape: translate one source sentence into one target language. In practice, users may ask for one target language, for several related languages at once, or for translations conditioned on examples. This paper studies prompt scope and demonstration selection as experimental variables for local LLM machine translation. We evaluate English-to-Romance and English-to-Germanic translation on the full FLORES devtest split for nine official European Union languages. We compare three local instruction-tuned LLMs, \texttt{llama3.2:3b}, \texttt{mistral:latest}, and \texttt{qwen2.5:14b}, against dedicated MT baselines from OPUS-MT and NLLB-200. We test zero-shot prompting and $k=5$ few-shot prompting with random, lexical-similarity, and embedding-similarity demonstration selection. We also compare single-target prompts with JSON-formatted family-scope prompts that request all languages in a family at once. Results show that dedicated MT systems remain strongest overall, especially for Germanic languages. Few-shot prompting helps \texttt{mistral:latest} and \texttt{qwen2.5:14b}, but hurts \texttt{llama3.2:3b}; embedding retrieval is best on average for the stronger LLMs, but its advantage over random and lexical examples is modest. Family-scope prompting is feasible for stronger local LLMs but exposes structured-output failures in smaller models. These findings motivate evaluating LLM translation not only by language pair and metric, but also by prompt scope, retrieval strategy, and multi-target compliance.
\end{abstract}

\section{Introduction}
\label{sec:introduction}

Machine translation (MT) evaluation traditionally treats translation as a language-pair task: a system is asked to translate from one source language into one target language, and the output is compared against reference translations. This framing remains appropriate for dedicated MT systems, but it does not fully capture how large language models (LLMs) are used as translation tools. LLM users can request one translation, several translations, or a set of related target languages in a single instruction. They can also provide examples, style constraints, and structured-output requirements. These extra degrees of freedom make prompt design part of the translation system itself.

This paper asks whether prompt scope should be treated as an explicit MT evaluation variable. We focus on local LLMs that can run on consumer hardware rather than on proprietary high-end systems. This setting matters because local deployment is attractive for privacy, cost control, and offline use, but smaller local models may be more sensitive to prompt length, example formatting, and structured-output requirements. We evaluate two prompt scopes. In the \emph{single-target} scope, the model translates English into one target language at a time. In the \emph{family-scope} setting, the model is asked to translate the same English sentence into all languages in a related family, returning a JSON object keyed by FLORES language codes.

We also study few-shot example selection. Prior work has shown that LLMs can be sensitive to in-context examples, but it is not obvious whether semantically similar examples improve translation more than random examples, especially in a multi-target prompt. We therefore compare zero-shot prompting with $k=5$ random examples, lexical-similarity examples, and embedding-similarity examples. Few-shot demonstrations are drawn from FLORES dev and evaluation is performed on FLORES devtest, avoiding leakage from the test split.

Our empirical study covers English-to-Romance and English-to-Germanic translation for nine official EU languages: French, Spanish, Italian, Portuguese, Romanian, German, Dutch, Danish, and Swedish. We evaluate three local instruction-tuned LLMs and compare them against dedicated MT baselines from OPUS-MT/MarianMT and NLLB-200. We report sacreBLEU BLEU, chrF++, COMET, paired bootstrap confidence intervals for key comparisons, sentence-level retrieval diagnostics, and family-scope compliance metrics such as target coverage and complete-output rate.

The results show that prompt scope and demonstration strategy materially affect local LLM translation. Dedicated MT systems remain strongest overall, but local LLMs are competitive for some Romance targets. Few-shot prompting helps \texttt{mistral:latest} and \texttt{qwen2.5:14b}, while degrading \texttt{llama3.2:3b}. Embedding-based retrieval is best on average for stronger models, but its margin over random and lexical retrieval is smaller than the differences between models. Family-scope prompting is reliable for stronger LLMs but can fail severely for smaller models, especially in the Germanic setting. The contribution of this paper is therefore not a new translation model, but an evaluation design and empirical analysis showing that prompt scope and structured-output compliance should be reported when LLMs are used for MT.

\section{Related Work}
\label{sec:related-work}

\paragraph{Machine translation benchmarks and metrics.}
FLORES-200 was introduced with No Language Left Behind (NLLB) as a multilingual evaluation benchmark covering many translation directions and professionally translated evaluation data \cite{nllb2022}. We use FLORES because it provides a consistent dev/devtest split across all target languages in our study. For automatic evaluation, BLEU remains widely reported, but score reproducibility depends on tokenization and implementation details; SacreBLEU was proposed to standardize BLEU reporting \cite{post2018sacrebleu}. We additionally report chrF++ \cite{popovic2015chrf,popovic2017chrfpp}, which is often useful for morphologically varied languages, and COMET, a learned metric designed to better correlate with human judgments \cite{rei2020comet,rei2022comet22}.

\paragraph{LLMs for machine translation.}
Recent work has evaluated GPT-style models as translation systems and shown that LLMs can be strong translators in high-resource settings while still exhibiting weaknesses relative to dedicated MT systems \cite{hendyetal2023gptmt,zhangetal2023promptingmt}. These studies motivate treating prompting and evaluation conditions carefully. Our work differs by focusing on local, consumer-hardware LLMs and by making prompt scope part of the experimental design.

\paragraph{In-context learning and example selection.}
In-context learning performance can depend strongly on which examples are placed in the prompt \cite{brown2020language,liu2022makes}. Retrieval-based example selection has been studied for language tasks, including selecting examples similar to the test input. For MT, however, an example can help by providing semantic analogies, terminology, output style, or simply the expected task format. We therefore include random demonstrations as a control, lexical retrieval as a lightweight baseline, and multilingual embedding retrieval as a semantic retrieval condition.

\paragraph{Terminology, lexical resources, and domain knowledge in MT.}
Terminology and lexical-resource translation are relevant to our setting because they isolate a problem that also appears in prompt-based LLM translation: translation quality often depends on whether the system receives enough semantic, lexical, or domain-specific evidence to choose the right target expression. Earlier work therefore provides a useful precedent for studying not only the translation model, but also the auxiliary knowledge made available to it. McCrae et al. combine semantic similarity with MT for cross-lingual knowledge graph translation, while Arcan et al. show that translating terminological expressions in knowledge bases is difficult without domain context and can benefit from adaptation or injected terminology \cite{mccrae2017linking,arcan2017terminological}. Related work has also used multi-way or multilingual NMT to infer dictionary entries and translate lexical resources, including closely related language settings and under-resourced WordNet gloss translation \cite{arcan2019inferring,chakravarthi2019wordnet}. Other studies examine whether rule-based linguistic, terminology, and named-entity knowledge can improve under-resourced NMT systems \cite{torregrosa2019leveraging,torregrosa2020aspects}. We do not inject terminology dictionaries or domain resources directly; instead, we test an analogous prompt-time question: whether retrieved demonstrations and related target languages provide useful semantic or lexical cues for local LLMs. This connection is why these works belong in the related work, and it also motivates our empirical caution: additional knowledge may help, but its benefit must be measured against random examples and zero-shot prompts rather than assumed.

\paragraph{Semantic ambiguity and translation context.}
Recent multimodal MT work also argues that translation quality can depend on whether additional context helps disambiguate the source. Hatami et al. use semantic diversity estimates to decide when visual information is likely to improve translation \cite{hatami2024enhancing}. Although our experiments are text-only, our sentence-level retrieval analysis is motivated by a similar question: when does additional contextual evidence help, and can a similarity score predict that benefit?

\paragraph{Structured and multi-target LLM outputs.}
LLMs are often used not only to generate text but to produce structured outputs such as JSON. In translation, requesting multiple target languages in one response can reduce orchestration cost but creates new failure modes: omitted targets, wrong-language outputs, prompt echoes, or malformed structures. Our family-scope setting evaluates this practical dimension directly by measuring both translation quality and target coverage.

\section{Methodology}
\label{sec:methodology}

\subsection{Task Definition}

Given an English source sentence $x$ and a set of target languages $T$, the model must produce translations $y_t$ for each $t \in T$. In the single-target setting, $|T|=1$ and each prompt requests exactly one translation. In the family-scope setting, $T$ is the set of languages in one family. For Romance, $T=\{\text{French},\text{Spanish},\text{Italian},\text{Portuguese},\text{Romanian}\}$; for Germanic, $T=\{\text{German},\text{Dutch},\text{Danish},\text{Swedish}\}$. Family-scope outputs are requested as JSON objects keyed by FLORES language codes.

\subsection{Prompt Conditions}

We evaluate four prompting conditions for both scopes. The zero-shot condition gives only the translation instruction and source sentence. The random few-shot condition prepends $k=5$ demonstration translations selected uniformly from the demonstration pool. The lexical retrieval condition selects the five FLORES dev examples with highest cosine similarity under bag-of-token source representations. The embedding retrieval condition selects the five examples with highest cosine similarity under multilingual sentence embeddings. All few-shot examples are selected from FLORES dev; no evaluation examples from FLORES devtest are used as demonstrations.

\subsection{Single-Target and Family-Scope Prompts}

Single-target prompts ask for one target language and instruct the model to return only the translation. Family-scope prompts ask for all languages in the family in a single JSON object. For few-shot family prompts, each demonstration contains the English source sentence and aligned translations for all targets in the same family. This design allows us to compare whether examples selected for the English source sentence transfer across all requested target languages.

\subsection{Evaluation Metrics}

For translation quality, we report corpus BLEU and chrF++ computed with SacreBLEU, and sentence-level COMET means computed with \texttt{Unbabel/wmt22-comet-da}. For paired comparisons between the best local LLM and the chrF++-best MT baseline, we use paired bootstrap resampling over matching FLORES source identifiers and report 95\% confidence intervals and two-sided $p$ values for chrF++ and COMET. For family-scope prompting, quality metrics are computed over produced target translations. We additionally report coverage, complete-output count, missing-target count, and prompt echo count.

\section{Experimental Setup}
\label{sec:experimental-setup}

\subsection{Data}

The main dataset is FLORES-200. Demonstrations are drawn from FLORES dev and evaluation is performed on the full FLORES devtest split. The devtest split contains 1012 English source sentences. We evaluate official EU languages from two families: Romance (French, Spanish, Italian, Portuguese, Romanian) and Germanic (German, Dutch, Danish, Swedish).

\subsection{Models}

The local LLMs are \texttt{llama3.2:3b}, \texttt{mistral:latest}, and \texttt{qwen2.5:14b}, run through Ollama with temperature 0. The dedicated MT baselines are \texttt{facebook/nllb-200-distilled-600M}, \texttt{facebook/nllb-200-distilled-1.3B}, and OPUS-MT/MarianMT models \cite{nllb2022,tiedemann2020opusmt,junczysdowmunt2018marian}. The MT baselines are evaluated only in the single-target setting, because they are designed as dedicated translation systems rather than structured multi-target instruction followers.

\subsection{Reproducibility}

Task files are deterministic and contain source identifiers, target language codes, references, prompt condition, prompt scope, and few-shot metadata. Few-shot source audits verify that demonstrations come from FLORES dev rather than FLORES devtest. Generation outputs are stored as JSONL files and evaluation tables are generated as CSV and TeX/Markdown summaries. The code and generated artifacts are organized so that single-target translation quality, family-scope compliance, COMET scoring, bootstrap significance, and sentence-level retrieval analysis can be reproduced independently.

\section{Results}
\label{sec:results}

We report the complete FLORES devtest outcomes for two official EU language families. Romance contains French, Spanish, Italian, Portuguese, and Romanian; Germanic contains German, Dutch, Danish, and Swedish. For each family we evaluate (i) single-target prompting, where each target language is translated separately, and (ii) family-scope prompting, where the model produces all translations for the family in one JSON-formatted response. Single-target prompting is the primary translation-quality setting; family-scope prompting is a combined translation and structured-output compliance setting.

All few-shot runs use $k=5$ demonstrations retrieved from FLORES dev, while evaluation is performed on FLORES devtest. We compare zero-shot prompting with random demonstrations, lexical-similarity retrieval, and embedding-similarity retrieval. Tables report corpus chrF++, corpus BLEU, and sentence-level COMET means. Compact baseline and diagnostic tables use slash-separated metric triples where noted in the caption.

\subsection{Romance languages}
\label{sec:results-romance}

Table~\ref{tab:romance-single-full} gives the full single-target Romance results for all local LLMs and prompting conditions. The strongest local model is usually \texttt{qwen2.5:14b}, especially for French, Spanish, Italian, and Portuguese. Romanian is the main exception, where \texttt{mistral:latest} with lexical examples gives the strongest local chrF++ score. Few-shot retrieval is not uniformly helpful: it consistently improves \texttt{mistral:latest} and \texttt{qwen2.5:14b} relative to their zero-shot variants, but it does not improve \texttt{llama3.2:3b}.

\begin{center}
\small
\setlength{\tabcolsep}{6pt}
\begin{longtable}{lllrrr}
\caption{Romance single-target local LLM outcomes on FLORES devtest. Few-shot conditions use $k=5$ examples from FLORES dev.}
\label{tab:romance-single-full}\\
\toprule
Target & Model & Condition & chrF++ & BLEU & COMET \\
\midrule
\endfirsthead
\caption[]{Romance single-target local LLM outcomes on FLORES devtest (continued).}\\
\toprule
Target & Model & Condition & chrF++ & BLEU & COMET \\
\midrule
\endhead
fra & \texttt{llama3.2:3b} & zero-shot & 60.55 & 36.53 & 0.8446 \\
fra & \texttt{llama3.2:3b} & random $k=5$ & 60.21 & 36.21 & 0.8436 \\
fra & \texttt{llama3.2:3b} & lexical $k=5$ & 60.23 & 36.31 & 0.8448 \\
fra & \texttt{llama3.2:3b} & embedding $k=5$ & 60.15 & 36.12 & 0.8438 \\
\addlinespace[0.2em]
fra & \texttt{mistral:latest} & zero-shot & 58.04 & 26.76 & 0.7992 \\
fra & \texttt{mistral:latest} & random $k=5$ & 61.68 & 37.19 & 0.8525 \\
fra & \texttt{mistral:latest} & lexical $k=5$ & 61.52 & 37.23 & 0.8517 \\
fra & \texttt{mistral:latest} & embedding $k=5$ & 61.39 & 37.38 & 0.8539 \\
\addlinespace[0.2em]
fra & \texttt{qwen2.5:14b} & zero-shot & 63.77 & 38.46 & 0.8514 \\
fra & \texttt{qwen2.5:14b} & random $k=5$ & 64.31 & 41.60 & 0.8649 \\
fra & \texttt{qwen2.5:14b} & lexical $k=5$ & 64.27 & 41.26 & 0.8642 \\
fra & \texttt{qwen2.5:14b} & embedding $k=5$ & 64.27 & 41.09 & 0.8672 \\
\addlinespace
spa & \texttt{llama3.2:3b} & zero-shot & 49.85 & 22.32 & 0.8368 \\
spa & \texttt{llama3.2:3b} & random $k=5$ & 49.36 & 21.92 & 0.8357 \\
spa & \texttt{llama3.2:3b} & lexical $k=5$ & 49.32 & 21.92 & 0.8349 \\
spa & \texttt{llama3.2:3b} & embedding $k=5$ & 49.79 & 22.33 & 0.8390 \\
\addlinespace[0.2em]
spa & \texttt{mistral:latest} & zero-shot & 47.05 & 15.01 & 0.7624 \\
spa & \texttt{mistral:latest} & random $k=5$ & 50.51 & 23.66 & 0.8415 \\
spa & \texttt{mistral:latest} & lexical $k=5$ & 50.59 & 23.81 & 0.8427 \\
spa & \texttt{mistral:latest} & embedding $k=5$ & 50.73 & 24.02 & 0.8448 \\
\addlinespace[0.2em]
spa & \texttt{qwen2.5:14b} & zero-shot & 51.43 & 24.02 & 0.8340 \\
spa & \texttt{qwen2.5:14b} & random $k=5$ & 52.27 & 26.12 & 0.8479 \\
spa & \texttt{qwen2.5:14b} & lexical $k=5$ & 51.99 & 25.72 & 0.8479 \\
spa & \texttt{qwen2.5:14b} & embedding $k=5$ & 52.59 & 26.53 & 0.8521 \\
\addlinespace
ita & \texttt{llama3.2:3b} & zero-shot & 50.14 & 22.40 & 0.8440 \\
ita & \texttt{llama3.2:3b} & random $k=5$ & 49.96 & 21.93 & 0.8472 \\
ita & \texttt{llama3.2:3b} & lexical $k=5$ & 50.01 & 22.08 & 0.8467 \\
ita & \texttt{llama3.2:3b} & embedding $k=5$ & 49.98 & 22.47 & 0.8452 \\
\addlinespace[0.2em]
ita & \texttt{mistral:latest} & zero-shot & 48.49 & 15.74 & 0.7906 \\
ita & \texttt{mistral:latest} & random $k=5$ & 51.11 & 22.92 & 0.8514 \\
ita & \texttt{mistral:latest} & lexical $k=5$ & 51.35 & 23.67 & 0.8559 \\
ita & \texttt{mistral:latest} & embedding $k=5$ & 51.70 & 24.13 & 0.8537 \\
\addlinespace[0.2em]
ita & \texttt{qwen2.5:14b} & zero-shot & 51.74 & 21.37 & 0.8414 \\
ita & \texttt{qwen2.5:14b} & random $k=5$ & 52.55 & 24.32 & 0.8621 \\
ita & \texttt{qwen2.5:14b} & lexical $k=5$ & 52.52 & 24.31 & 0.8594 \\
ita & \texttt{qwen2.5:14b} & embedding $k=5$ & 52.78 & 25.03 & 0.8646 \\
\addlinespace
por & \texttt{llama3.2:3b} & zero-shot & 62.69 & 38.91 & 0.8703 \\
por & \texttt{llama3.2:3b} & random $k=5$ & 61.44 & 37.23 & 0.8682 \\
por & \texttt{llama3.2:3b} & lexical $k=5$ & 61.45 & 37.12 & 0.8679 \\
por & \texttt{llama3.2:3b} & embedding $k=5$ & 61.45 & 37.01 & 0.8696 \\
\addlinespace[0.2em]
por & \texttt{mistral:latest} & zero-shot & 55.68 & 21.73 & 0.7805 \\
por & \texttt{mistral:latest} & random $k=5$ & 61.62 & 37.09 & 0.8698 \\
por & \texttt{mistral:latest} & lexical $k=5$ & 61.70 & 37.41 & 0.8692 \\
por & \texttt{mistral:latest} & embedding $k=5$ & 62.13 & 37.91 & 0.8728 \\
\addlinespace[0.2em]
por & \texttt{qwen2.5:14b} & zero-shot & 65.04 & 40.55 & 0.8722 \\
por & \texttt{qwen2.5:14b} & random $k=5$ & 65.44 & 42.65 & 0.8791 \\
por & \texttt{qwen2.5:14b} & lexical $k=5$ & 65.69 & 43.07 & 0.8807 \\
por & \texttt{qwen2.5:14b} & embedding $k=5$ & 65.08 & 41.53 & 0.8796 \\
\addlinespace
ron & \texttt{llama3.2:3b} & zero-shot & 51.12 & 24.77 & 0.8182 \\
ron & \texttt{llama3.2:3b} & random $k=5$ & 50.00 & 24.09 & 0.8119 \\
ron & \texttt{llama3.2:3b} & lexical $k=5$ & 50.27 & 24.06 & 0.8127 \\
ron & \texttt{llama3.2:3b} & embedding $k=5$ & 50.20 & 23.99 & 0.8168 \\
\addlinespace[0.2em]
ron & \texttt{mistral:latest} & zero-shot & 48.28 & 15.61 & 0.7408 \\
ron & \texttt{mistral:latest} & random $k=5$ & 52.24 & 25.21 & 0.8401 \\
ron & \texttt{mistral:latest} & lexical $k=5$ & 53.25 & 27.40 & 0.8423 \\
ron & \texttt{mistral:latest} & embedding $k=5$ & 53.13 & 26.68 & 0.8456 \\
\addlinespace[0.2em]
ron & \texttt{qwen2.5:14b} & zero-shot & 49.08 & 18.46 & 0.7632 \\
ron & \texttt{qwen2.5:14b} & random $k=5$ & 50.00 & 20.84 & 0.7877 \\
ron & \texttt{qwen2.5:14b} & lexical $k=5$ & 50.88 & 22.41 & 0.7925 \\
ron & \texttt{qwen2.5:14b} & embedding $k=5$ & 50.99 & 22.50 & 0.7997 \\
\bottomrule
\end{longtable}
\end{center}

Table~\ref{tab:romance-mt-full} reports the dedicated MT baselines for the same Romance targets. The MT systems remain strong, with the chrF++-best systems being OPUS-MT for French and Romanian and NLLB-1.3B for Spanish, Italian, and Portuguese. The local LLMs are closest to MT on Spanish and Portuguese, while Romanian remains substantially harder.

\begin{center}
\small
\setlength{\tabcolsep}{6pt}
\begin{longtable}{lllll}
\caption{Romance dedicated MT baseline outcomes. Each metric cell reports chrF++/BLEU/COMET.}
\label{tab:romance-mt-full}\\
\toprule
Target & OPUS-MT & NLLB-600M & NLLB-1.3B & Best chrF++ \\
\midrule
fra & 67.95/47.81/0.8616 & 65.24/45.08/0.8586 & 67.06/47.48/0.8710 & OPUS-MT \\
spa & 52.83/26.68/0.8499 & 51.91/26.08/0.8492 & 52.97/27.30/0.8569 & NLLB-1.3B \\
ita & 54.75/27.53/0.8521 & 53.55/26.87/0.8598 & 55.40/29.34/0.8740 & NLLB-1.3B \\
por & 37.04/7.37/0.7468 & 65.92/44.92/0.8780 & 66.87/46.33/0.8873 & NLLB-1.3B \\
ron & 58.18/32.90/0.8686 & 56.39/31.90/0.8733 & 58.06/33.80/0.8908 & OPUS-MT \\
\bottomrule
\end{longtable}
\end{center}

Family-scope Romance results are shown at the target level in Table~\ref{tab:romance-family-target-full} and as compliance-aware family averages in Table~\ref{tab:romance-family-full}. In this setting the model must produce all five Romance translations in one structured response. \texttt{qwen2.5:14b} gives the best average quality, while \texttt{mistral:latest} and \texttt{qwen2.5:14b} both maintain high coverage. \texttt{llama3.2:3b} is less reliable: even though its zero-shot family coverage is above 0.91, coverage drops under few-shot family prompting, showing that additional context can make structured multi-target output harder for the smaller model.

\begin{center}
\small
\setlength{\tabcolsep}{6pt}
\begin{longtable}{lllrrr}
\caption{Romance family-scope target-level quality for outputs produced when all Romance targets are requested together in one JSON response.}
\label{tab:romance-family-target-full}\\
\toprule
Target & Model & Condition & chrF++ & BLEU & COMET \\
\midrule
\endfirsthead
\caption[]{Romance family-scope target-level quality (continued).}\\
\toprule
Target & Model & Condition & chrF++ & BLEU & COMET \\
\midrule
\endhead
fra & \texttt{llama3.2:3b} & zero-shot & 48.31 & 26.01 & 0.8075 \\
fra & \texttt{llama3.2:3b} & random $k=5$ & 52.67 & 30.41 & 0.8341 \\
fra & \texttt{llama3.2:3b} & lexical $k=5$ & 53.50 & 30.85 & 0.8331 \\
fra & \texttt{llama3.2:3b} & embedding $k=5$ & 53.44 & 30.77 & 0.8374 \\
\addlinespace[0.2em]
fra & \texttt{mistral:latest} & zero-shot & 60.30 & 37.15 & 0.8486 \\
fra & \texttt{mistral:latest} & random $k=5$ & 61.93 & 38.79 & 0.8584 \\
fra & \texttt{mistral:latest} & lexical $k=5$ & 61.84 & 38.80 & 0.8574 \\
fra & \texttt{mistral:latest} & embedding $k=5$ & 61.90 & 38.83 & 0.8573 \\
\addlinespace[0.2em]
fra & \texttt{qwen2.5:14b} & zero-shot & 64.12 & 42.47 & 0.8730 \\
fra & \texttt{qwen2.5:14b} & random $k=5$ & 64.54 & 42.94 & 0.8792 \\
fra & \texttt{qwen2.5:14b} & lexical $k=5$ & 64.66 & 43.00 & 0.8787 \\
fra & \texttt{qwen2.5:14b} & embedding $k=5$ & 64.71 & 43.04 & 0.8789 \\
\addlinespace
spa & \texttt{llama3.2:3b} & zero-shot & 39.78 & 15.40 & 0.8175 \\
spa & \texttt{llama3.2:3b} & random $k=5$ & 43.25 & 18.09 & 0.8368 \\
spa & \texttt{llama3.2:3b} & lexical $k=5$ & 43.92 & 18.37 & 0.8369 \\
spa & \texttt{llama3.2:3b} & embedding $k=5$ & 44.02 & 18.68 & 0.8358 \\
\addlinespace[0.2em]
spa & \texttt{mistral:latest} & zero-shot & 49.04 & 21.99 & 0.8395 \\
spa & \texttt{mistral:latest} & random $k=5$ & 50.48 & 23.47 & 0.8506 \\
spa & \texttt{mistral:latest} & lexical $k=5$ & 50.37 & 23.19 & 0.8493 \\
spa & \texttt{mistral:latest} & embedding $k=5$ & 50.59 & 23.80 & 0.8493 \\
\addlinespace[0.2em]
spa & \texttt{qwen2.5:14b} & zero-shot & 51.87 & 25.12 & 0.8636 \\
spa & \texttt{qwen2.5:14b} & random $k=5$ & 52.19 & 25.51 & 0.8666 \\
spa & \texttt{qwen2.5:14b} & lexical $k=5$ & 52.14 & 25.50 & 0.8663 \\
spa & \texttt{qwen2.5:14b} & embedding $k=5$ & 52.04 & 25.44 & 0.8661 \\
\addlinespace
ita & \texttt{llama3.2:3b} & zero-shot & 40.21 & 15.61 & 0.8259 \\
ita & \texttt{llama3.2:3b} & random $k=5$ & 43.71 & 18.49 & 0.8464 \\
ita & \texttt{llama3.2:3b} & lexical $k=5$ & 44.43 & 18.83 & 0.8475 \\
ita & \texttt{llama3.2:3b} & embedding $k=5$ & 44.40 & 18.68 & 0.8481 \\
\addlinespace[0.2em]
ita & \texttt{mistral:latest} & zero-shot & 49.63 & 21.83 & 0.8528 \\
ita & \texttt{mistral:latest} & random $k=5$ & 51.09 & 23.36 & 0.8637 \\
ita & \texttt{mistral:latest} & lexical $k=5$ & 51.06 & 23.17 & 0.8637 \\
ita & \texttt{mistral:latest} & embedding $k=5$ & 51.25 & 23.57 & 0.8632 \\
\addlinespace[0.2em]
ita & \texttt{qwen2.5:14b} & zero-shot & 52.61 & 25.16 & 0.8763 \\
ita & \texttt{qwen2.5:14b} & random $k=5$ & 52.57 & 24.81 & 0.8796 \\
ita & \texttt{qwen2.5:14b} & lexical $k=5$ & 52.59 & 25.09 & 0.8781 \\
ita & \texttt{qwen2.5:14b} & embedding $k=5$ & 52.90 & 25.29 & 0.8802 \\
\addlinespace
por & \texttt{llama3.2:3b} & zero-shot & 48.52 & 25.30 & 0.8459 \\
por & \texttt{llama3.2:3b} & random $k=5$ & 52.71 & 29.73 & 0.8686 \\
por & \texttt{llama3.2:3b} & lexical $k=5$ & 53.79 & 30.43 & 0.8682 \\
por & \texttt{llama3.2:3b} & embedding $k=5$ & 53.36 & 29.73 & 0.8667 \\
\addlinespace[0.2em]
por & \texttt{mistral:latest} & zero-shot & 58.99 & 34.15 & 0.8632 \\
por & \texttt{mistral:latest} & random $k=5$ & 59.80 & 35.06 & 0.8717 \\
por & \texttt{mistral:latest} & lexical $k=5$ & 59.82 & 35.03 & 0.8714 \\
por & \texttt{mistral:latest} & embedding $k=5$ & 60.09 & 35.27 & 0.8721 \\
\addlinespace[0.2em]
por & \texttt{qwen2.5:14b} & zero-shot & 64.22 & 41.35 & 0.8927 \\
por & \texttt{qwen2.5:14b} & random $k=5$ & 64.38 & 41.54 & 0.8945 \\
por & \texttt{qwen2.5:14b} & lexical $k=5$ & 64.31 & 41.42 & 0.8946 \\
por & \texttt{qwen2.5:14b} & embedding $k=5$ & 64.62 & 41.80 & 0.8952 \\
\addlinespace
ron & \texttt{llama3.2:3b} & zero-shot & 38.74 & 15.32 & 0.7573 \\
ron & \texttt{llama3.2:3b} & random $k=5$ & 42.76 & 18.16 & 0.8018 \\
ron & \texttt{llama3.2:3b} & lexical $k=5$ & 44.19 & 19.15 & 0.8025 \\
ron & \texttt{llama3.2:3b} & embedding $k=5$ & 43.94 & 19.02 & 0.8051 \\
\addlinespace[0.2em]
ron & \texttt{mistral:latest} & zero-shot & 50.51 & 24.29 & 0.8214 \\
ron & \texttt{mistral:latest} & random $k=5$ & 51.83 & 25.73 & 0.8350 \\
ron & \texttt{mistral:latest} & lexical $k=5$ & 52.07 & 25.89 & 0.8397 \\
ron & \texttt{mistral:latest} & embedding $k=5$ & 51.98 & 25.96 & 0.8415 \\
\addlinespace[0.2em]
ron & \texttt{qwen2.5:14b} & zero-shot & 51.42 & 25.29 & 0.8207 \\
ron & \texttt{qwen2.5:14b} & random $k=5$ & 51.75 & 25.80 & 0.8328 \\
ron & \texttt{qwen2.5:14b} & lexical $k=5$ & 51.99 & 25.93 & 0.8340 \\
ron & \texttt{qwen2.5:14b} & embedding $k=5$ & 51.97 & 26.00 & 0.8375 \\
\bottomrule
\end{longtable}
\end{center}

\begin{center}
\small
\setlength{\tabcolsep}{6pt}
\begin{longtable}{llrrr}
\caption{Romance family-scope JSON outcomes. Quality metrics are averaged across Romance target languages. Complete items are out of 1012 source sentences; missing targets are summed across French, Spanish, Italian, Portuguese, and Romanian.}
\label{tab:romance-family-full}\\
\toprule
Model & Condition & chrF++ & BLEU & COMET \\
\midrule
\multicolumn{5}{l}{\textit{Quality on produced translations}} \\
\texttt{llama3.2:3b} & zero-shot & 43.11 & 19.53 & 0.8108 \\
\texttt{llama3.2:3b} & random $k=5$ & 47.02 & 22.98 & 0.8375 \\
\texttt{llama3.2:3b} & lexical $k=5$ & 47.97 & 23.52 & 0.8376 \\
\texttt{llama3.2:3b} & embedding $k=5$ & 47.83 & 23.37 & 0.8386 \\
\addlinespace
\texttt{mistral:latest} & zero-shot & 53.69 & 27.88 & 0.8451 \\
\texttt{mistral:latest} & random $k=5$ & 55.02 & 29.28 & 0.8559 \\
\texttt{mistral:latest} & lexical $k=5$ & 55.03 & 29.21 & 0.8563 \\
\texttt{mistral:latest} & embedding $k=5$ & 55.16 & 29.49 & 0.8567 \\
\addlinespace
\texttt{qwen2.5:14b} & zero-shot & 56.85 & 31.88 & 0.8652 \\
\texttt{qwen2.5:14b} & random $k=5$ & 57.09 & 32.12 & 0.8705 \\
\texttt{qwen2.5:14b} & lexical $k=5$ & 57.14 & 32.19 & 0.8703 \\
\texttt{qwen2.5:14b} & embedding $k=5$ & 57.25 & 32.31 & 0.8716 \\
\midrule
Model & Condition & Coverage & Complete & Missing \\
\midrule
\multicolumn{5}{l}{\textit{Structured-output compliance}} \\
\texttt{llama3.2:3b} & zero-shot & 0.914 & 925/1012 & 434 \\
\texttt{llama3.2:3b} & random $k=5$ & 0.863 & 873/1012 & 695 \\
\texttt{llama3.2:3b} & lexical $k=5$ & 0.876 & 887/1012 & 625 \\
\texttt{llama3.2:3b} & embedding $k=5$ & 0.870 & 880/1012 & 660 \\
\addlinespace
\texttt{mistral:latest} & zero-shot & 0.994 & 1006/1012 & 30 \\
\texttt{mistral:latest} & random $k=5$ & 0.993 & 1005/1012 & 35 \\
\texttt{mistral:latest} & lexical $k=5$ & 0.991 & 1003/1012 & 44 \\
\texttt{mistral:latest} & embedding $k=5$ & 0.994 & 1006/1012 & 30 \\
\addlinespace
\texttt{qwen2.5:14b} & zero-shot & 0.985 & 997/1012 & 75 \\
\texttt{qwen2.5:14b} & random $k=5$ & 0.986 & 998/1012 & 70 \\
\texttt{qwen2.5:14b} & lexical $k=5$ & 0.986 & 997/1012 & 71 \\
\texttt{qwen2.5:14b} & embedding $k=5$ & 0.986 & 996/1012 & 72 \\
\bottomrule
\end{longtable}
\end{center}
\subsection{Germanic languages}
\label{sec:results-germanic}

Table~\ref{tab:germanic-single-full} gives the full single-target Germanic results. The same model-dependent few-shot pattern appears: \texttt{mistral:latest} and \texttt{qwen2.5:14b} improve with retrieved demonstrations, while \texttt{llama3.2:3b} generally degrades. Germanic targets show larger gaps to dedicated MT systems than Romance targets, particularly for Danish and Swedish.

\begin{center}
\small
\setlength{\tabcolsep}{6pt}
\begin{longtable}{lllrrr}
\caption{Germanic single-target local LLM outcomes on FLORES devtest. Each row reports chrF++/BLEU/COMET for one target, model, and prompt condition. Few-shot conditions use $k=5$ examples from FLORES dev.}
\label{tab:germanic-single-full}\\
\toprule
Target & Model & Condition & chrF++ & BLEU & COMET \\
\midrule
\endfirsthead
\caption[]{Germanic single-target local LLM outcomes on FLORES devtest (continued).}\\
\toprule
Target & Model & Condition & chrF++ & BLEU & COMET \\
\midrule
\endhead
deu & \texttt{llama3.2:3b} & zero-shot & 54.32 & 26.58 & 0.8299 \\
deu & \texttt{llama3.2:3b} & random $k=5$ & 53.92 & 26.13 & 0.8292 \\
deu & \texttt{llama3.2:3b} & lexical $k=5$ & 53.47 & 25.86 & 0.8274 \\
deu & \texttt{llama3.2:3b} & embedding $k=5$ & 53.52 & 25.93 & 0.8267 \\
\addlinespace[0.2em]
deu & \texttt{mistral:latest} & zero-shot & 50.28 & 15.71 & 0.7414 \\
deu & \texttt{mistral:latest} & random $k=5$ & 53.82 & 24.24 & 0.8299 \\
deu & \texttt{mistral:latest} & lexical $k=5$ & 53.77 & 24.39 & 0.8309 \\
deu & \texttt{mistral:latest} & embedding $k=5$ & 53.70 & 24.16 & 0.8336 \\
\addlinespace[0.2em]
deu & \texttt{qwen2.5:14b} & zero-shot & 54.99 & 24.58 & 0.8195 \\
deu & \texttt{qwen2.5:14b} & random $k=5$ & 55.49 & 26.80 & 0.8403 \\
deu & \texttt{qwen2.5:14b} & lexical $k=5$ & 55.48 & 27.24 & 0.8410 \\
deu & \texttt{qwen2.5:14b} & embedding $k=5$ & 55.95 & 27.95 & 0.8422 \\
\addlinespace
nld & \texttt{llama3.2:3b} & zero-shot & 48.46 & 19.23 & 0.8285 \\
nld & \texttt{llama3.2:3b} & random $k=5$ & 48.36 & 19.39 & 0.8256 \\
nld & \texttt{llama3.2:3b} & lexical $k=5$ & 48.17 & 18.93 & 0.8266 \\
nld & \texttt{llama3.2:3b} & embedding $k=5$ & 48.19 & 19.18 & 0.8278 \\
\addlinespace[0.2em]
nld & \texttt{mistral:latest} & zero-shot & 45.31 & 11.13 & 0.7446 \\
nld & \texttt{mistral:latest} & random $k=5$ & 48.62 & 18.33 & 0.8350 \\
nld & \texttt{mistral:latest} & lexical $k=5$ & 48.88 & 18.58 & 0.8362 \\
nld & \texttt{mistral:latest} & embedding $k=5$ & 49.08 & 19.11 & 0.8406 \\
\addlinespace[0.2em]
nld & \texttt{qwen2.5:14b} & zero-shot & 48.41 & 16.41 & 0.8145 \\
nld & \texttt{qwen2.5:14b} & random $k=5$ & 49.27 & 18.98 & 0.8377 \\
nld & \texttt{qwen2.5:14b} & lexical $k=5$ & 49.08 & 18.51 & 0.8343 \\
nld & \texttt{qwen2.5:14b} & embedding $k=5$ & 49.35 & 18.94 & 0.8416 \\
\addlinespace
dan & \texttt{llama3.2:3b} & zero-shot & 52.16 & 25.62 & 0.7889 \\
dan & \texttt{llama3.2:3b} & random $k=5$ & 51.60 & 25.12 & 0.7890 \\
dan & \texttt{llama3.2:3b} & lexical $k=5$ & 51.77 & 25.45 & 0.7909 \\
dan & \texttt{llama3.2:3b} & embedding $k=5$ & 51.74 & 25.46 & 0.7969 \\
\addlinespace[0.2em]
dan & \texttt{mistral:latest} & zero-shot & 49.30 & 15.05 & 0.7093 \\
dan & \texttt{mistral:latest} & random $k=5$ & 55.69 & 28.98 & 0.8478 \\
dan & \texttt{mistral:latest} & lexical $k=5$ & 55.87 & 29.21 & 0.8520 \\
dan & \texttt{mistral:latest} & embedding $k=5$ & 56.22 & 30.05 & 0.8572 \\
\addlinespace[0.2em]
dan & \texttt{qwen2.5:14b} & zero-shot & 49.53 & 17.71 & 0.7323 \\
dan & \texttt{qwen2.5:14b} & random $k=5$ & 50.46 & 19.71 & 0.7656 \\
dan & \texttt{qwen2.5:14b} & lexical $k=5$ & 50.78 & 20.02 & 0.7668 \\
dan & \texttt{qwen2.5:14b} & embedding $k=5$ & 51.30 & 20.89 & 0.7739 \\
\addlinespace
swe & \texttt{llama3.2:3b} & zero-shot & 56.06 & 30.25 & 0.8369 \\
swe & \texttt{llama3.2:3b} & random $k=5$ & 55.28 & 29.29 & 0.8351 \\
swe & \texttt{llama3.2:3b} & lexical $k=5$ & 55.26 & 29.65 & 0.8358 \\
swe & \texttt{llama3.2:3b} & embedding $k=5$ & 55.26 & 29.79 & 0.8399 \\
\addlinespace[0.2em]
swe & \texttt{mistral:latest} & zero-shot & 50.80 & 16.51 & 0.7495 \\
swe & \texttt{mistral:latest} & random $k=5$ & 56.18 & 28.33 & 0.8555 \\
swe & \texttt{mistral:latest} & lexical $k=5$ & 56.23 & 28.58 & 0.8599 \\
swe & \texttt{mistral:latest} & embedding $k=5$ & 56.59 & 29.09 & 0.8625 \\
\addlinespace[0.2em]
swe & \texttt{qwen2.5:14b} & zero-shot & 50.20 & 17.85 & 0.7625 \\
swe & \texttt{qwen2.5:14b} & random $k=5$ & 52.18 & 21.70 & 0.7947 \\
swe & \texttt{qwen2.5:14b} & lexical $k=5$ & 52.47 & 22.13 & 0.7977 \\
swe & \texttt{qwen2.5:14b} & embedding $k=5$ & 53.34 & 23.49 & 0.8077 \\
\bottomrule
\end{longtable}
\end{center}

Table~\ref{tab:germanic-mt-full} reports the Germanic MT baselines. OPUS-MT is the strongest baseline by chrF++ and BLEU for all four Germanic targets in this experiment, with especially large margins for Danish and Swedish. These results show that strong dedicated MT models remain difficult to match with small local instruction-tuned LLMs.

\begin{center}
\small
\setlength{\tabcolsep}{6pt}
\begin{longtable}{lllll}
\caption{Germanic dedicated MT baseline outcomes. Each metric cell reports chrF++/BLEU/COMET.}
\label{tab:germanic-mt-full}\\
\toprule
Target & OPUS-MT & NLLB-600M & NLLB-1.3B & Best chrF++ \\
\midrule
deu & 61.11/36.04/0.8467 & 57.38/32.27/0.8421 & 59.32/35.23/0.8615 & OPUS-MT \\
nld & 53.80/25.01/0.8499 & 52.18/24.42/0.8509 & 53.39/25.54/0.8656 & OPUS-MT \\
dan & 66.25/43.37/0.8822 & 61.43/38.16/0.8780 & 62.89/40.36/0.8935 & OPUS-MT \\
swe & 65.99/43.70/0.8896 & 60.64/37.01/0.8753 & 62.53/39.81/0.8935 & OPUS-MT \\
\bottomrule
\end{longtable}
\end{center}

Tables~\ref{tab:germanic-family-target-full} and~\ref{tab:germanic-family-full} show the Germanic family-scope results. This is the clearest compliance stress test in the paper. \texttt{mistral:latest} and \texttt{qwen2.5:14b} produce mostly complete JSON outputs, with embedding coverage of 0.980 and 0.972 respectively. \texttt{llama3.2:3b}, however, fails the family-scope format frequently: its Germanic zero-shot coverage is only 0.343, and even embedding retrieval reaches only 0.493. For this model, the family-scope result is therefore primarily a structured-output failure, not only a translation-quality failure.

\begin{center}
\small
\setlength{\tabcolsep}{6pt}
\begin{longtable}{lllrrr}
\caption{Germanic family-scope target-level quality. Each row reports chrF++/BLEU/COMET for one target, model, and prompt condition when all Germanic targets are requested together in one JSON response. Compliance for these same runs is summarized in Table~\ref{tab:germanic-family-full}.}
\label{tab:germanic-family-target-full}\\
\toprule
Target & Model & Condition & chrF++ & BLEU & COMET \\
\midrule
\endfirsthead
\caption[]{Germanic family-scope target-level quality (continued).}\\
\toprule
Target & Model & Condition & chrF++ & BLEU & COMET \\
\midrule
\endhead
deu & \texttt{llama3.2:3b} & zero-shot & 15.45 & 1.05 & 0.8058 \\
deu & \texttt{llama3.2:3b} & random $k=5$ & 26.84 & 7.26 & 0.8278 \\
deu & \texttt{llama3.2:3b} & lexical $k=5$ & 23.41 & 5.25 & 0.8224 \\
deu & \texttt{llama3.2:3b} & embedding $k=5$ & 26.96 & 7.39 & 0.8316 \\
\addlinespace[0.2em]
deu & \texttt{mistral:latest} & zero-shot & 51.67 & 24.23 & 0.8277 \\
deu & \texttt{mistral:latest} & random $k=5$ & 53.29 & 26.08 & 0.8414 \\
deu & \texttt{mistral:latest} & lexical $k=5$ & 53.05 & 25.92 & 0.8420 \\
deu & \texttt{mistral:latest} & embedding $k=5$ & 53.67 & 26.32 & 0.8439 \\
\addlinespace[0.2em]
deu & \texttt{qwen2.5:14b} & zero-shot & 55.27 & 29.04 & 0.8608 \\
deu & \texttt{qwen2.5:14b} & random $k=5$ & 55.73 & 29.35 & 0.8646 \\
deu & \texttt{qwen2.5:14b} & lexical $k=5$ & 55.70 & 29.08 & 0.8655 \\
deu & \texttt{qwen2.5:14b} & embedding $k=5$ & 56.15 & 29.82 & 0.8652 \\
\addlinespace
nld & \texttt{llama3.2:3b} & zero-shot & 13.81 & 0.78 & 0.7943 \\
nld & \texttt{llama3.2:3b} & random $k=5$ & 23.72 & 5.20 & 0.8266 \\
nld & \texttt{llama3.2:3b} & lexical $k=5$ & 20.90 & 3.77 & 0.8214 \\
nld & \texttt{llama3.2:3b} & embedding $k=5$ & 23.99 & 5.35 & 0.8274 \\
\addlinespace[0.2em]
nld & \texttt{mistral:latest} & zero-shot & 47.01 & 18.20 & 0.8309 \\
nld & \texttt{mistral:latest} & random $k=5$ & 48.04 & 18.97 & 0.8428 \\
nld & \texttt{mistral:latest} & lexical $k=5$ & 48.20 & 19.28 & 0.8438 \\
nld & \texttt{mistral:latest} & embedding $k=5$ & 48.36 & 19.26 & 0.8429 \\
\addlinespace[0.2em]
nld & \texttt{qwen2.5:14b} & zero-shot & 48.87 & 20.61 & 0.8558 \\
nld & \texttt{qwen2.5:14b} & random $k=5$ & 48.97 & 20.36 & 0.8599 \\
nld & \texttt{qwen2.5:14b} & lexical $k=5$ & 49.16 & 20.41 & 0.8625 \\
nld & \texttt{qwen2.5:14b} & embedding $k=5$ & 49.29 & 20.38 & 0.8621 \\
\addlinespace
dan & \texttt{llama3.2:3b} & zero-shot & 13.52 & 0.76 & 0.7305 \\
dan & \texttt{llama3.2:3b} & random $k=5$ & 24.80 & 6.05 & 0.7737 \\
dan & \texttt{llama3.2:3b} & lexical $k=5$ & 21.55 & 4.44 & 0.7707 \\
dan & \texttt{llama3.2:3b} & embedding $k=5$ & 24.67 & 6.23 & 0.7740 \\
\addlinespace[0.2em]
dan & \texttt{mistral:latest} & zero-shot & 52.01 & 25.30 & 0.8366 \\
dan & \texttt{mistral:latest} & random $k=5$ & 53.20 & 26.37 & 0.8528 \\
dan & \texttt{mistral:latest} & lexical $k=5$ & 53.27 & 26.95 & 0.8521 \\
dan & \texttt{mistral:latest} & embedding $k=5$ & 53.78 & 27.41 & 0.8537 \\
\addlinespace[0.2em]
dan & \texttt{qwen2.5:14b} & zero-shot & 52.13 & 26.42 & 0.8250 \\
dan & \texttt{qwen2.5:14b} & random $k=5$ & 52.62 & 26.82 & 0.8359 \\
dan & \texttt{qwen2.5:14b} & lexical $k=5$ & 53.17 & 27.23 & 0.8374 \\
dan & \texttt{qwen2.5:14b} & embedding $k=5$ & 53.54 & 27.74 & 0.8359 \\
\addlinespace
swe & \texttt{llama3.2:3b} & zero-shot & 15.58 & 1.12 & 0.8188 \\
swe & \texttt{llama3.2:3b} & random $k=5$ & 26.73 & 7.71 & 0.8425 \\
swe & \texttt{llama3.2:3b} & lexical $k=5$ & 23.30 & 5.53 & 0.8353 \\
swe & \texttt{llama3.2:3b} & embedding $k=5$ & 26.45 & 7.60 & 0.8396 \\
\addlinespace[0.2em]
swe & \texttt{mistral:latest} & zero-shot & 53.43 & 26.94 & 0.8586 \\
swe & \texttt{mistral:latest} & random $k=5$ & 54.34 & 28.30 & 0.8691 \\
swe & \texttt{mistral:latest} & lexical $k=5$ & 54.06 & 27.98 & 0.8676 \\
swe & \texttt{mistral:latest} & embedding $k=5$ & 54.80 & 28.69 & 0.8688 \\
\addlinespace[0.2em]
swe & \texttt{qwen2.5:14b} & zero-shot & 52.92 & 26.83 & 0.8429 \\
swe & \texttt{qwen2.5:14b} & random $k=5$ & 53.06 & 26.76 & 0.8500 \\
swe & \texttt{qwen2.5:14b} & lexical $k=5$ & 53.17 & 26.75 & 0.8475 \\
swe & \texttt{qwen2.5:14b} & embedding $k=5$ & 53.54 & 27.17 & 0.8462 \\
\bottomrule
\end{longtable}
\end{center}

\begin{center}
\small
\setlength{\tabcolsep}{6pt}
\begin{longtable}{llrrr}
\caption{Germanic family-scope JSON outcomes. Quality metrics are averaged across Germanic target languages. Complete items are out of 1012 source sentences; missing targets are summed across German, Dutch, Danish, and Swedish.}
\label{tab:germanic-family-full}\\
\toprule
Model & Condition & chrF++ & BLEU & COMET \\
\midrule
\multicolumn{5}{l}{\textit{Quality on produced translations}} \\
\midrule
\texttt{llama3.2:3b} & zero-shot & 14.59 & 0.93 & 0.7873 \\
\texttt{llama3.2:3b} & random $k=5$ & 25.53 & 6.55 & 0.8177 \\
\texttt{llama3.2:3b} & lexical $k=5$ & 22.29 & 4.75 & 0.8124 \\
\texttt{llama3.2:3b} & embedding $k=5$ & 25.52 & 6.64 & 0.8181 \\
\addlinespace
\texttt{mistral:latest} & zero-shot & 51.03 & 23.67 & 0.8385 \\
\texttt{mistral:latest} & random $k=5$ & 52.22 & 24.93 & 0.8515 \\
\texttt{mistral:latest} & lexical $k=5$ & 52.14 & 25.03 & 0.8514 \\
\texttt{mistral:latest} & embedding $k=5$ & 52.65 & 25.42 & 0.8523 \\
\addlinespace
\texttt{qwen2.5:14b} & zero-shot & 52.30 & 25.73 & 0.8461 \\
\texttt{qwen2.5:14b} & random $k=5$ & 52.59 & 25.82 & 0.8526 \\
\texttt{qwen2.5:14b} & lexical $k=5$ & 52.80 & 25.87 & 0.8532 \\
\texttt{qwen2.5:14b} & embedding $k=5$ & 53.13 & 26.28 & 0.8524 \\
\midrule
Model & Condition & Coverage & Complete & Missing \\
\midrule
\multicolumn{5}{l}{\textit{Structured-output compliance}} \\
\midrule
\texttt{llama3.2:3b} & zero-shot & 0.343 & 347/1012 & 2660 \\
\texttt{llama3.2:3b} & random $k=5$ & 0.487 & 493/1012 & 2076 \\
\texttt{llama3.2:3b} & lexical $k=5$ & 0.437 & 442/1012 & 2278 \\
\texttt{llama3.2:3b} & embedding $k=5$ & 0.493 & 499/1012 & 2052 \\
\addlinespace
\texttt{mistral:latest} & zero-shot & 0.987 & 999/1012 & 52 \\
\texttt{mistral:latest} & random $k=5$ & 0.975 & 987/1012 & 100 \\
\texttt{mistral:latest} & lexical $k=5$ & 0.975 & 987/1012 & 100 \\
\texttt{mistral:latest} & embedding $k=5$ & 0.980 & 992/1012 & 80 \\
\addlinespace
\texttt{qwen2.5:14b} & zero-shot & 0.960 & 972/1012 & 160 \\
\texttt{qwen2.5:14b} & random $k=5$ & 0.965 & 977/1012 & 140 \\
\texttt{qwen2.5:14b} & lexical $k=5$ & 0.968 & 980/1012 & 128 \\
\texttt{qwen2.5:14b} & embedding $k=5$ & 0.972 & 984/1012 & 112 \\
\bottomrule
\end{longtable}
\end{center}

\subsection{Few-shot retrieval effects}
\label{sec:results-retrieval}

Table~\ref{tab:retrieval-deltas-all} aggregates few-shot deltas over all nine target languages. The main finding is that retrieval is model-dependent. \texttt{llama3.2:3b} loses chrF++ under all retrieval settings. \texttt{mistral:latest} gains strongly under all retrieval settings, with embedding retrieval giving the largest average gain (+4.60 chrF++, +11.03 BLEU, +0.0941 COMET). \texttt{qwen2.5:14b} also improves across all targets, with embedding retrieval again strongest on average (+1.27 chrF++, +3.17 BLEU, +0.0264 COMET).

\begin{center}
\small
\setlength{\tabcolsep}{6pt}
\begin{longtable}{llll}
\caption{Mean few-shot deltas over all nine target languages relative to zero-shot. Each cell reports $\Delta$chrF++/$\Delta$BLEU/$\Delta$COMET.}
\label{tab:retrieval-deltas-all}\\
\toprule
Model & random & lexical & embedding \\
\midrule
\texttt{llama3.2:3b} & -0.58/-0.59/-0.0014 & -0.60/-0.58/-0.0012 & -0.56/-0.48/+0.0008 \\
\texttt{mistral:latest} & +4.25/+10.30/+0.0895 & +4.44/+10.78/+0.0914 & +4.60/+11.03/+0.0941 \\
\texttt{qwen2.5:14b} & +0.87/+2.59/+0.0210 & +0.99/+2.81/+0.0215 & +1.27/+3.17/+0.0264 \\
\bottomrule
\end{longtable}
\end{center}

Embedding retrieval is therefore the best average retrieval method for the two models that benefit from demonstrations. However, the advantage over lexical and random retrieval is smaller than the difference between models. This indicates that the ability to use demonstrations is the dominant factor; semantic retrieval helps, but it does not rescue a model that is already brittle under longer prompts.

\subsection{Comparison with dedicated MT systems}
\label{sec:results-mt-comparison}

Table~\ref{tab:best-llm-vs-mt-significance} gives paired significance tests for the best local LLM condition against the chrF++-best MT baseline per target. The best local LLM remains below this MT baseline on chrF++ for all targets. The average chrF++ gap is smaller for Romance languages (-2.55) than for Germanic languages (-7.26). Spanish is the closest target: the chrF++ difference between \texttt{qwen2.5:14b} with embedding retrieval and NLLB-1.3B is not significant ($p=0.520$). COMET narrows some differences: French slightly favors the local LLM over the chrF++-best MT baseline in mean COMET but not significantly, and German shows a non-significant COMET difference despite a significant chrF++ gap.

\begin{center}
\small
\setlength{\tabcolsep}{6pt}
\begin{longtable}{lllrrrrr}
\caption{Paired comparison of the best local LLM condition and the chrF++-best dedicated MT baseline per target. Deltas are local LLM minus MT baseline.}
\label{tab:best-llm-vs-mt-significance}\\
\toprule
Target & Best local LLM & chrF-best & $\Delta$chrF & 95\% CI & $p_{\mathrm{chrF}}$ & $\Delta$COMET & $p_{\mathrm{COMET}}$ \\
\midrule
fra & \texttt{qwen2.5:14b} random & OPUS-MT & -3.63 & [-4.53, -2.97] & 0.000 & +0.0033 & 0.200 \\
spa & \texttt{qwen2.5:14b} embedding & NLLB-1.3B & -0.38 & [-0.69, 0.40] & 0.520 & -0.0049 & 0.072 \\
ita & \texttt{qwen2.5:14b} embedding & NLLB-1.3B & -2.62 & [-3.03, -1.76] & 0.000 & -0.0094 & 0.000 \\
por & \texttt{qwen2.5:14b} lexical & NLLB-1.3B & -1.18 & [-1.66, -0.07] & 0.020 & -0.0066 & 0.008 \\
ron & \texttt{mistral:latest} lexical & OPUS-MT & -4.93 & [-5.46, -4.02] & 0.000 & -0.0263 & 0.000 \\
deu & \texttt{qwen2.5:14b} embedding & OPUS-MT & -5.16 & [-5.72, -4.22] & 0.000 & -0.0046 & 0.236 \\
nld & \texttt{qwen2.5:14b} embedding & OPUS-MT & -4.45 & [-4.71, -3.52] & 0.000 & -0.0083 & 0.004 \\
dan & \texttt{mistral:latest} embedding & OPUS-MT & -10.03 & [-10.77, -9.18] & 0.000 & -0.0250 & 0.000 \\
swe & \texttt{mistral:latest} embedding & OPUS-MT & -9.40 & [-9.75, -8.21] & 0.000 & -0.0271 & 0.000 \\
\bottomrule
\end{longtable}
\end{center}

\subsection{Sentence-level retrieval analysis}
\label{sec:results-sentence-level}

Table~\ref{tab:sentence-retrieval-full} reports sentence-level retrieval diagnostics. The correlation between the demonstration similarity score and the sentence-level chrF gain is close to zero in almost all settings. This is important: even when embedding retrieval is the best average condition, higher embedding similarity does not strongly predict that a specific sentence will improve. Retrieval gain is therefore not simply a monotonic function of the similarity score used for selecting demonstrations.

\begin{center}
\small
\setlength{\tabcolsep}{6pt}
\begin{longtable}{lllll}
\caption{Sentence-level retrieval diagnostics. Each cell reports mean $\Delta$chrF/mean $\Delta$BLEU/positive-rate/correlation between demonstration similarity score and sentence-level chrF gain.}
\label{tab:sentence-retrieval-full}\\
\toprule
Romance \\
\midrule
Model & random & lexical & embedding \\
\midrule
\texttt{llama3.2:3b} & -0.69/-0.88/0.42/-0.000 & -0.62/-0.78/0.43/-0.003 & -0.64/-0.81/0.43/+0.015 \\
\texttt{mistral:latest} & +3.70/+7.45/0.67/+0.019 & +3.80/+7.53/0.68/+0.005 & +3.97/+7.73/0.68/+0.025 \\
\texttt{qwen2.5:14b} & +0.44/+0.65/0.42/-0.023 & +0.60/+0.89/0.44/-0.006 & +0.67/+0.93/0.46/+0.002 \\
\midrule
Germanic \\
\midrule
Model & random & lexical & embedding \\
\midrule
\texttt{llama3.2:3b} & -0.50/-0.44/0.44/+0.004 & -0.64/-0.43/0.45/-0.019 & -0.62/-0.45/0.45/-0.009 \\
\texttt{mistral:latest} & +4.63/+8.29/0.70/+0.025 & +4.76/+8.36/0.72/+0.004 & +4.86/+8.74/0.72/+0.019 \\
\texttt{qwen2.5:14b} & +0.88/+1.16/0.47/+0.001 & +0.99/+1.19/0.49/-0.022 & +1.37/+1.65/0.52/+0.023 \\
\bottomrule
\end{longtable}
\end{center}

Length effects are more visible but still model-specific. For \texttt{mistral:latest}, retrieval gains are larger on longer sentences in both families. With embedding retrieval, Romance gains increase from +2.53 chrF on the shortest length quartile to +4.92 on the longest, and Germanic gains increase from +3.30 to +5.36. For \texttt{qwen2.5:14b}, the length pattern is weaker and sometimes reversed. The sentence-level analysis therefore supports a cautious interpretation: retrieved examples help when the model can use them, but neither lexical nor embedding similarity score is sufficient by itself to predict sentence-level gains.

\subsection{Summary of findings}
\label{sec:results-summary}

The complete outcomes support five findings. First, dedicated MT systems remain the strongest local translation systems on FLORES devtest, especially for Germanic languages. Second, local LLMs are competitive for some Romance targets, particularly Spanish and Portuguese, but they do not consistently match MT baselines. Third, few-shot prompting is beneficial for \texttt{mistral:latest} and \texttt{qwen2.5:14b}, while it hurts \texttt{llama3.2:3b}; this makes model capacity and instruction-following ability central to few-shot MT prompting. Fourth, embedding retrieval is the best average retrieval method for the models that benefit from examples, but its advantage over lexical and random examples is modest. Fifth, family-scope prompting should be reported as both translation and compliance: larger local LLMs can produce mostly complete structured multi-target outputs, whereas smaller models can omit required languages, especially in the Germanic family setting.

These results motivate treating prompt scope as an experimental variable in LLM-based MT evaluation. Single-target prompting measures translation quality most cleanly; family-scope prompting tests whether a local LLM can function as a structured multi-target translation engine; and retrieval experiments reveal whether demonstrations help because of semantic relevance or because a given model is generally capable of using in-context examples.

\section{Discussion and Limitations}
\label{sec:discussion}

The results show that prompt scope is not a superficial formatting choice in LLM-based machine translation. A single-target prompt and a family-scope prompt ask for different behaviours: the former tests translation into one language, while the latter also tests whether the model can coordinate several target languages and obey a structured output schema. This distinction matters in practice. Family-scope prompting is attractive because it can produce several translations in one generation call, but it also introduces failure modes that standard single-target MT evaluation would miss, including omitted languages, incomplete JSON objects, and cross-target interference. The Germanic family results for \texttt{llama3.2:3b} make this particularly clear: the model can produce reasonable single-target translations, but frequently fails to return all required languages when asked for the whole family at once.

The few-shot results also caution against a simple ``more similar examples are better'' interpretation. Embedding retrieval gives the best average gains for \texttt{mistral:latest} and \texttt{qwen2.5:14b}, but the sentence-level correlations between retrieval similarity score and translation improvement are close to zero. This suggests that demonstrations help through several mechanisms at once: they can clarify the task format, anchor the output style, expose target-language conventions, and occasionally provide useful semantic parallels. For smaller models, however, the same additional context can become a burden rather than a benefit. Random demonstrations are therefore not merely a weak baseline; they are a necessary control for separating the effect of example relevance from the more general effect of showing the model the expected translation pattern.

The comparison with dedicated MT systems provides an important boundary for the claims. Local LLMs are useful and competitive in selected settings, especially for some Romance targets, but the strongest dedicated MT baselines remain better overall in chrF++ and BLEU, with the largest gaps on Germanic languages. COMET narrows some differences and occasionally gives a less severe view of the LLM outputs, but it does not overturn the main conclusion. The contribution of this study is therefore not that small local LLMs replace dedicated MT systems. Rather, it is that prompt scope, demonstration selection, and structured-output compliance substantially affect how local LLM translation should be evaluated and reported.

Several limitations remain. The study uses English as the only source language and covers two European language families. The patterns may differ for non-English sources, lower-resource languages, non-Latin scripts, morphologically richer languages, or domain-specific corpora. We also evaluate a fixed set of local model checkpoints through one inference stack and with temperature set to 0; other decoding settings, context lengths, quantizations, or instruction templates may change both translation quality and output compliance. The evaluation is automatic: BLEU, chrF++, and COMET provide complementary signals, but they do not replace human adequacy and fluency assessment. In the family-scope setting, quality scores are computed over produced translations, so they must always be interpreted together with coverage and missing-output counts. A model that produces fluent translations for only part of the requested family is not equivalent to a model that reliably produces every target.

The practical implication is that local LLM translation should be deployed with validation rather than assumed reliability. Local inference is attractive for privacy-sensitive settings because source text need not be sent to a remote API, but local execution does not by itself guarantee correctness. Missing outputs, wrong-language responses, prompt echoes, and fluent mistranslations can create real risks in legal, medical, administrative, or safety-critical contexts. For multi-target translation, structured-output checks should be treated as part of the system, not as optional post-processing. Users should preserve source text, model metadata, prompts, and validation logs, and should avoid relying on automatic translations as the sole basis for high-stakes decisions.

\section{Conclusion}
\label{sec:conclusion}

This paper evaluated local LLM machine translation under two prompt scopes and three few-shot demonstration strategies. On FLORES English-to-Romance and English-to-Germanic translation, dedicated MT systems remain the strongest overall baselines, while local LLMs are competitive for selected Romance targets. Few-shot prompting helps stronger local LLMs but hurts the smallest model, and embedding-based retrieval is best on average without being universally predictive at the sentence level. Family-scope prompting exposes a different capability: the ability to produce complete structured multi-target translations. Larger local LLMs mostly satisfy this requirement, while smaller models can fail by omitting languages. These findings support evaluating LLM translation as a combination of translation quality, prompt scope, demonstration selection, and output compliance.

\bibliographystyle{plain}
\bibliography{references}

\end{document}